\pdfoutput=1
\documentclass[11pt]{article}

\usepackage{style/naacl2021}

\usepackage{times}
\usepackage{array}

\newif\ifcomment\commentfalse

\usepackage{framed}
\usepackage{mdwlist}
\usepackage{fnpct}
\usepackage{latexsym}
\usepackage{colortbl}
\usepackage{xcolor}
\usepackage{nicefrac}
\usepackage{booktabs}
\usepackage{amsfonts}
\usepackage[T1]{fontenc}
\usepackage{bold-extra}
\usepackage{amsmath}
\usepackage{amssymb}
\usepackage{bm}
\usepackage{graphicx}
\usepackage{mathtools}
\usepackage{microtype}
\usepackage{multirow}
\usepackage{multicol}
\usepackage{latexsym,comment}

\graphicspath{ {2020_tacl_factcheck/images/} }

\newcommand{\gem}[1]{\mbox{\textsc{gem}}}
\newcommand{\abr}[1]{\textsc{#1}}

\newcommand{\smallemaillink}[2]{ {\small \href{mailto://#2}{\texttt{#1}}}}

\newcommand{\hidetext}[1]{}
\newcommand{\ignore}[1]{}

\ifcomment
\newcommand{\pinaforecomment}[3]{\colorbox{#1}{\parbox{.8\linewidth}{#2: #3}}}
\else
\newcommand{\pinaforecomment}[3]{}
\fi

\newcommand{\smallurl}[1]{ \begin{tiny}\url{#1}\end{tiny}}

\definecolor{lightblue}{HTML}{3cc7ea}
\definecolor{CUgold}{HTML}{CFB87C}
\definecolor{grey}{rgb}{0.95,0.95,0.95}
\definecolor{ceil}{rgb}{0.57, 0.63, 0.81}
\definecolor{UMDred}{HTML}{ed1c24}
\definecolor{UMDyellow}{HTML}{ffc20e}

\usepackage{sistyle}
\SIthousandsep{,}

\usepackage[utf8]{inputenc}

\newcommand{\fever}{\textsc{fever}}

\newcommand{\hardfever}{\textsc{hard fever}}
\newcommand{\fmt}{\textsc{fm2}}
\newcommand{\easy}{\textsc{easy}}
\newcommand{\hard}{\textsc{hard}}

\newcommand{\bigfmt}{\textsc{FoolMeTwice}}

\newcommand{\figfile}[1]{2020_tacl_factcheck/figures/#1}

\title{Fool Me Twice: Entailment from Wikipedia Gamification}

\author{
Julian Martin Eisenschlos,
Bhuwan Dhingra, \\
\textbf{Jannis Bulian, Benjamin B{\"o}rschinger } \\
Google Research \\ 
\\
\small{\texttt{\{}}\smallemaillink{eisenjulian}{eisenjulian@google.com},
\smallemaillink{bdhingra}{bdhingra@google.com}, 
\smallemaillink{jbulian}{jbulian@google.com}, 
\smallemaillink{bboerschinger}{bboerschinger@google.com}\small{\texttt{\}}} \\
\small{\texttt{@google.com}}
\And
Jordan Boyd-Graber\thanks{\; Work completed while a Visiting Research Scientist at Google.} \\
\\
\abr{cs}, iSchool, \abr{umiacs}, \abr{lsc} \\
University of Maryland\\
\smallemaillink{jbg@umiacs.umd.edu}{jbg@umiacs.umd.edu} \\\\
}

\begin{document}

\maketitle

\begin{abstract}
We release \bigfmt~(\fmt~for short), a large dataset of challenging entailment pairs collected through a fun multi-player game.
Gamification encourages adversarial examples, drastically lowering the number of examples that can be solved using ``shortcuts'' compared to other entailment datasets.
Players are presented with two tasks. 
The first task asks the player to write a plausible claim based on the evidence from a Wikipedia page.  
The second one shows two plausible claims written by other players, one of which is false, and the goal is to identify it before the time runs out.
Players ``pay'' to see clues retrieved from the evidence pool: the more evidence the player needs, the harder the claim.
%
%
Game-play between motivated players leads to diverse strategies for crafting claims, such as temporal inference and diverting to unrelated evidence, and results in higher quality data for the entailment and evidence retrieval tasks.
We open source the dataset and game code.\footnote{\url{https://github.com/google-research/fool-me-twice}}
\end{abstract}

\section{Introducing a Game of Challenging Claims}
\label{sec:into}

\begin{figure}[ht]
    \centering
    \includegraphics[width=.8\linewidth]{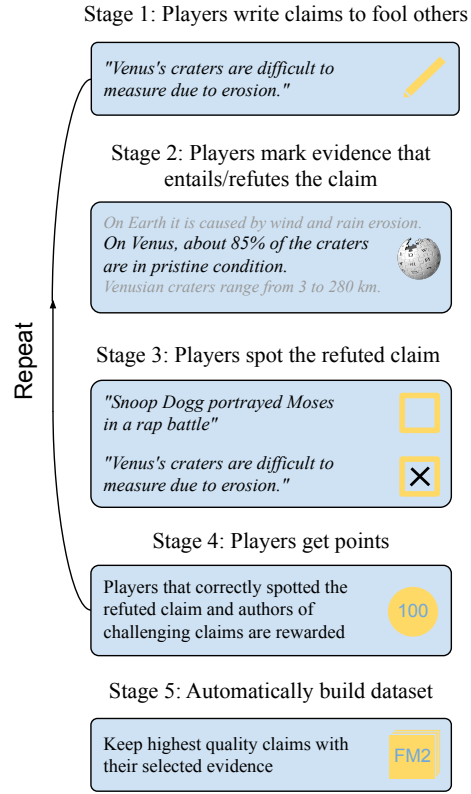}
    \caption{Overview of the data generation pipeline. In stages 1 to 4, players write challenging claims either entailed or refuted by evidence from Wikipedia (Section~\ref{sec:human-authored}). They are then tasked to spot the refuted claim among a group (Section~\ref{sec:what-is-hard}). The claims and evidence are available for download.}
    \label{fig:pipeline}
\end{figure}

Given a statement---and a large collection of textual knowledge---how
do you find evidence that shows a reader that the statement is true or
false?
This problem takes on multiple forms in the natural language
processing (\abr{nlp}) community.
Given only a single statement and a single sentence, this decision
process is called \emph{recognizing textual entailment}~\cite[\abr{rte}]{dagan-10} or \emph{natural language inference}~\cite[\abr{nli}]{snli, williams-etal-2018-broad}.
Given a single statement and a vast pool of possible evidence (e.g.,
all of Wikipedia), this problem is called
\emph{verification}~\cite{fever-18, jiang-etal-2020-hover}.

We review existing resources for the latter task in
Section~\ref{sec:related} and how they have spawned a vibrant
sub-community around related tasks.
However, these datasets fail to challenge modern \abr{nlp} models such
as \abr{bert}~\cite{devlin-19} or \abr{T5}~\cite{t5} that achieve
``super-human performance'' despite also exhibiting ``annotation artifacts'' that
hurt their generalization
potential~\cite{gururangan-etal-2018-annotation,
  tsuchiya-2018-performance}.
%
Our goal is twofold: (1) to build a new, challenging
dataset (statistics for \bigfmt{} in Table~\ref{tab:dataset-stats})
that tests models' ability to retrieve evidence and verify
claims and (2) to show that engineering the incentive structure of
data collection experiments can produce more accurate and realistic outcomes.

This dataset lends itself to automatic training and it characterizes
what factual errors humans can most easily detect and which are most
likely to fool them (Section~\ref{sec:what-is-hard}).
This is analogous to the creation of unsupported or refuted claims in the wild,
which are not random, but evolve as part of an information arms race~\cite{rid-20}.
Unlike previous datasets that rely on crowd-sourcing, we develop an
online game to create a platform where motivated authors can create
plausible sounding ``facts'' that other users must debunk.
%

Not only does this create more realistic claims---the best must
withstand human scrutiny---it also creates a way to better evaluate
the evidence that support or refute claims.
As we surface the evidence, humans use that evidence to decide which claims are
true or false; these signals can further improve our systems (Figure~\ref{fig:pipeline}).
We apply baseline models for retrieval and classification to our
dataset (Section~\ref{sec:models}) and examine how their ability to
detect wrong statements differs from humans'
(Section~\ref{sec:results}). 
%

\begin{table}
\begin{center}
\scalebox{0.8}{
\begin{tabular}{cccc>{\centering\arraybackslash}p{8mm}>{\centering\arraybackslash}p{12mm}}  
\toprule
               & Claims & Entailed  & Pages & \multicolumn{2}{c}{Avg. \# Tokens}\\
               &        & Proportion     &       & Claim & Evidence\\
\midrule
\textbf{Train} & 10,419  & 49.2\%   & 1,811  & 15           & 30              \\
\textbf{Dev}   & 1,169   & 51.0\%   & 209   & 15           & 31              \\
\textbf{Test}  & 1,380   & 49.4\%   & 234   & 15           & 31              \\
\midrule
\textbf{Total} & 12,968  & 49.4\%   & 2,254  & 15           & 30              \\
\bottomrule
\end{tabular}
}
\end{center}
\caption{Statistics of the \bigfmt~dataset. The train/dev/test split is based on disjoint Wikipedia pages. The number of tokens is an average
  value computed with a white-space tokenizer.  Our dataset is balanced between entailed (true) and not entailed (false) claims.}
\label{tab:dataset-stats}
\end{table}
%

\section{Related Work}
\label{sec:related}


Entailment is a key task in natural language understanding.
\newcite{dagan-10} describe it as an \abr{ai}-complete task: solve it,
and you can solve all of artificial intelligence.
Typically, entailment is presented as: given a premise (``Brooklyn is
the most populous of New York City's boroughs''), decide whether a
hypothesis (``Manhattan has more residents than Brooklyn'') is
entailed---supported---by the premise.
Even simple examples show the promise (and complexity) of this task.
To recognize that this hypothesis is contradicted, a model must: know
that Manhattan is a borough of New York, recognize that ``X is the most
populous borough'' entails ``X has more residents than any other borough'',
and correctly combine this knowledge to recognize the contradiction.

\subsection{Entailment and Retrieval Datasets}

Despite the promise of entailment, it has not been a silver bullet for
the \abr{nlp} community to solve artificial intelligence.
One possible explanation is highlighted by a line of work that shows existing
entailment datasets have artifacts.
\newcite{poliak-18} show entailment can often be solved by looking only at
the hypothesis, while \newcite{Wallace-19b} show that artifacts can
infect the premise as well.
This is especially common in the biggest datasets for \abr{nli} such
as \abr{snli} and \abr{mnli}~\cite{gururangan-etal-2018-annotation}.
While there are algorithmic solutions to addressing these issues~\cite{utama-20a}, many have turned to building better datasets.

Both \citet{bowman-etal-2020-new} and \citet{vania2020asking} propose
alternative methods for collecting entailment pairs from crowdworkers
and measure success via improvements in other general tasks via
transfer learning.
While the proposed methods prove to be ineffective for that goal, we
view \abr{nli} is as an important end task in itself (e.g., for
misinformation, \abr{qa}, dialogue, generation evaluation).
Hence, we argue that constructing challenging entailment datasets is
useful beyond just transfer
learning.

Like this paper, \newcite{nie-etal-2020-adversarial} focus on
adversarial entailment, but their authors only see a single piece of
evidence.
We expand this human-in-the-loop adversarial setting to include the essential
\emph{retrieval} component of fact verification.
Thus, authors have more strategies on hand; in addition to creating
challenging examples through paraphrasing, they can make it difficult
to \emph{find} relevant information in the first place or distract
with related---but distinct---information.

This is exactly the setting of a recent shared task, \fever{}~\cite[Fact Extraction and
VERification]{fever-18}, which creates a more general entailment
setting:
given a claim,
find relevant evidence from Wikipedia,
and determine whether the evidence has enough information to either
support or refute the claim.
This generalizes the entailment problem to a large, broadly accepted
set of premises (all sentences in Wikipedia) and adds an additional
retrieval step to find relevant evidence.

\fever{} has obvious connections to problems in education, journalism,
and information science.  Thus, it has caught the attention of a
subcommunity focused on building systems for \fever{} shared tasks.
Despite this excitement, \citet{schuster-etal-2019-towards} show that
\fever{} has many of the same issues as entailment datasets.
\fever{} has broad or nonsensical claims (Table~\ref{tab:howlers}) and
many of the claims are generated from the very first line of source
Wikipedia documents.
This is not just an artifact of crowd-sourcing; a more fundamental
problem is that there is no clear definition of what makes a good
\fever{} example.
To date, adversarial \fever{} example generation uses automatic rules to
increase their difficulty~\cite{thorne-19}.
To address these identified weaknesses, Sections~\ref{sec:human-authored}
and~\ref{sec:what-is-hard} define a game
where the claim writers have a clear objective of ``fooling'' other
human players.

\begin{table}[t!]
    \centering
    \begin{tabular}{c}
    \cellcolor[HTML]{C0C0C0} {\bf Supported} \\
    Woody Allen is a person. \\
    The Shining was directed. \\
    François de Belleforest wrote. \\
    \cellcolor[HTML]{C0C0C0} {\bf Not Enough Info} \\
    Lisa Kudrow was in a car. \\
    Tipper Gore was curated to Al Gore. \\
    International Relations includes animals. \\
    \cellcolor[HTML]{C0C0C0} {\bf Refuted} \\
    Tipper Gore was created in 1048. \\
    Alpha House is inspired by nobody. \\
    Toy Story is incapable of being a film. \\
    \end{tabular}
    \caption{Examples from \fever{},
    which separates entailment examples into three categories.  The crowdworkers who authored the examples often edit the first line of the Wikipedia article but not in ways that sound like a plausible hypothesis.  We develop a game to build more complex, challenging examples.  }
    \label{tab:howlers}
\end{table}

\subsection{Gamification for Data Collection}

Creating datasets through a fun interactive design is often called \emph{gamification}.
\citet{quizz-game} focus on multiple choice question answering in
technical domains such as medicine and rely on redundancy and
calibration questions to generate new knowledge.
The \abr{esp} game~\cite{esp-game} asks users to write labels for an
image that agree as much as possible with other players' labels.

Another well-known example is protein folding~\cite{Cooper2010}, an online game\footnote{\url{https://fold.it}} that tasks players
to twist and bend protein structures, often besting computer algorithms and driving biological innovations~\cite{Khatib2011}.

Crucially, these games are either individual or cooperative; in
contrast, \bigfmt{} exploits the adversarial nature of 
players fooling each other.
\bigfmt{} most closely resembles \textit{Balderdash}, a board game where players guess which definition of a word is legitimate that is  used
in information literacy courses~\cite{hays-17}.

In all cases, the intrinsic motivation driven by these games can lead
to better outcomes and fewer attempts to ``game'' the
system~\cite{motivations1, motivations2}.
Thus our approach constitutes a viable alternative to traditional
isolated labelling tasks in crowd-sourcing platforms,
where tying payment to
completing tasks sometimes hurts final results~\cite{payenough}.

\section{\bigfmt{} Game Mechanics}

This section outlines the two phases of the game: authoring claims (Section~\ref{sec:human-authored}) and voting on those claims (Section~\ref{sec:what-is-hard}).  While these sections present the game in its final form, this is the reflection of an iterative process.

 We first began with a paper version~\cite{nielsen-89} of the game, which showed that a time constraint made the game feel more fun and encouraged people to not read individual pieces of evidence too intently.  Without the timer, people tried to look for tiny clues in text that probably were not there~\cite{wilkinson-12}.  We then moved onto a version of the game presented via slides
 where we experimented with design choices such as the number
 of claims players distinguish between,
 and the number of evidence sentences they see while doing that.
Examples of the final web interface are shown in Appendix~\ref{app:game}.

\subsection{Crafting Challenging Claims}
\label{sec:human-authored}

\begin{table}
\centering
\resizebox{1.0\columnwidth}{!}{
\begin{tabular}{l|l}
\toprule													
Dataset split	&	Top Bigrams	by \abr{lmi} (highest predictive power first) \\
\midrule													
\fever~ Train & \textbf{is only}, \textbf{did not}, not a, was not, \textbf{incapable of}, \textbf{only a} \\
\fever~ Dev   & \textbf{is only}, \textbf{only a}, \textbf{incapable of}, is incapable, was only, \textbf{did not} \\
\midrule
\fmt~ Train   & the second, is a, was a, was the, is the, of his \\ 
\fmt~ Dev     & by a, on the, innocent iii, statue of, for his, pope innocen \\
\bottomrule
\end{tabular}
}
\caption{Top $6$ bigrams with the highest
  \abr{lmi}~\cite{schuster-etal-2019-towards} for \textsc{refutes} in
  each dataset and each split. Overlapping bigrams are bolded. Compared to
  \fever{}, \bigfmt~ contains fewer bigrams that ``give away'' the
  label on both the train and dev set.}
\label{tab:all_artefacts}
\end{table}

Our goal is to create a computer game that produces human-authored,
interesting, challenging claims paired with evidence that either
supports or refutes each claim.
One prerequisite for this is that claims avoid high lexical overlap with
the knowledge corpus. We thus need to encourage authors to craft
claims that cannot be trivially matched to evidence.
While this approach has been used for question
answering~\cite{wallace-19,bartolo-20}, which has a similar retrieval step, to
our knowledge it has not been applied to entailment or \fever{}.

We recruit users employed at Google, all proficient in English, to play-test the game.
At the beginning of each round, we ask each user to generate a true or
false statement.
We randomly choose a Wikipedia page as a knowledge source and ask them
to highlight one or two evidence spans that support (or refute) their claim.
They are instructed to write statements that would likely \emph{fool}
other players trying to determine the claim's veracity quickly and/or
without looking at the evidence that support the claim.
The reward system defined in the next section is built to be aligned
with this objective.

To help authors write hard claims, not entirely similar to the
evidence, we show the user what evidence a \abr{tf-idf} retrieval
system would select from the source
and highlight the words that help \abr{ir} systems select evidence.
This implicitly encourages them to craft the claims in a manner such
that overlap with the evidence is low (Section~\ref{sec:what-is-hard}).
We include screenshots of the user interface and more details about
our design choices in the appendix.
%
%
Because the players see evidence selected by our retrieval systems,
difficult claims for players are also challenging
for computers.
See Table~\ref{tab:all_artefacts} for a comparison on highly
predictive bigrams between \fever~ and \bigfmt~
(details about how these are computed are in the appendix).


\subsection{Spotting the Incorrect Statement}
\label{sec:what-is-hard}

In the game's second phase, players select the
incorrect statement from claims written by other players
(Table~\ref{tab:claims}).
To separate these two phases of the game, we refer to players in this
phase of the game as \emph{voters}.
If a voter can correctly answer quickly (e.g., through their own world
knowledge or artifacts), they get up to $120$ points, the
maximum possible.\footnote{Each voting task should take at maximum two minutes, and each point corresponds to a second.}
The author and voter split the points: any points the voter leaves ``on the
table'' go to the author.  Challenging claims reward the author with more
points but easy ones let the \emph{voter} increase their total.

We do not want to keep claims that are easy to identify as true or false.
If the average player can tell through artifacts or common sense that
a claim will not be supported, it is uninteresting as an entailment
example.
For example, if someone sees the claim ``Tipper Gore was born in
1048'' and remembers that Al Gore was the vice president of the United
States in the twentieth century, they can identify that this claim is
false.
We also want claims that require the voters to carefully read evidence from Wikipedia
(Table~\ref{tab:claims}).
Voters can ask for hints provided by our evidence selection system
(Section~\ref{sec:ir}).
For each piece of evidence shown, the number of points available to
the voters decreases, and points decrease as time progresses as well.

All possible outcomes provide useful information: correct and
incorrect choices, with and without evidence.
As mentioned before, if voters spot the wrong statement unaided, the claim
has underlying issues.
When a voter can spot the wrong claim with the help of a particular piece of evidence, then this is a clue
that the evidence (and the mechanism that selected it) is useful.

This allows us to specifically optimize for evidence that \emph{helps}
players better answer questions.
When voters go from confused to confident about the correct answer,
that is a signal that the evidence was effective.
When voters select an incorrect answer, that is a signal that the
evidence was not effective (or, indeed, misleading).

When voters need more time and evidence and are almost fooled
(i.e., nearly think a true statement is incorrect), this is a sign that the
statement is challenging for the human--computer team seeking to
verify entailment.
The statement must be convincingly written, consistent with voter's
world knowledge, \emph{and} also consistent with the evidence players
see.
Our game setting helps create conditions where these ``tricky''
examples can be crafted.

We use two heuristics to ensure quality claims.
First, we search for ``easy'' examples that were consistently
solved without inspecting the evidence -- however, we were not able to find any.
Next, we search for examples which are ``too difficult'' by
computing a \textit{maximum a posteriori} estimate of the Bernoulli distribution of correct and incorrect votes for each claim.
The prior distribution matches the overall accuracy of the dataset (80\% of votes are correct) and is equivalent to adding five pseudocounts (one wrong, four correct) for each question.
We use this smoothed estimate rather than the maximum likelihood estimate to account for claims lacking votes.
The expected value of that posterior given a Beta$(4,1)$ prior is~\cite{NIPS2012_cd00692c}:
\begin{align*}
\alpha & \sim{\rm Beta}\left(4, 1\right) \\
\alpha\,\mid\,C_i & \sim{\rm Beta}\left(4 + \sum_i C_i, 1 + \sum_i\left(1 - C_i\right)\right),    
\end{align*}
where $i$ sums over the votes, and $C_i$ is one if the vote was correct and zero otherwise.
We analyze all twenty-five claims below a $0.5$ threshold and identified three
incorrect examples which we subsequently removed.
%
%
%


\begin{table*}[ht]
\begin{scriptsize}
\begin{tabular}{p{1.35cm}p{6cm}p{6cm}r}
{\bf Topic} & 
\centering \textit{Iceland} &
\centering \textit{Elizabeth II} &
Time Left \\
{\bf Claim} & 
After ending its personal union with Denmark, Iceland was last invaded by the United Kingdom in Operation Fork.  &
After Elizabeth II's accession, she changed her house's name from "House of Saxe-Coburg and Gotha" to the "House of Windsor", rejecting the "House of Edinburgh".  &
3:00 \\
\toprule
\rowcolor[HTML]{C0C0C0} {\bf Retrieved \newline Evidence} &
The Danish-Icelandic Act of Union, an agreement with Denmark signed on 1 December 1918 and valid for 25 years, recognised Iceland as a fully sovereign and independent state in a personal union with Denmark.&
Philip suggested House of Edinburgh, after his ducal title. &
-0:30 \\
\rowcolor[HTML]{D0D0D0} (Revealed \newline Incrementally) &
Possession of Iceland passed from the Kingdom of Norway (872–1397) to the Kalmar Union in 1415, when the kingdoms of Norway, Denmark and Sweden were united.&
The Duke's uncle, Lord Mountbatten, advocated the name House of Mountbatten. &
-0:30 \\
\rowcolor[HTML]{E0E0E0} &
A month later, British armed forces conducted Operation Fork, the invasion and occupation of the country, violating Icelandic neutrality.&
The Duke complained, "I am the only man in the country not allowed to give his name to his own children." &
-0:30 \\
\rowcolor[HTML]{F0F0F0} &
Beginning on 20 May 1944, Icelanders voted in a four-day plebiscite on whether to terminate the personal union with Denmark, abolish the monarchy, and establish a republic.&
With Elizabeth's accession, it seemed probable the royal house would bear the Duke of Edinburgh's name, in line with the custom of a wife taking her husband's surname on marriage. &
-0:30 \\
& $\vdots$ & $\vdots$ & \\
\midrule
{\bf Gold \newline Evidence} &
After the German occupation of Denmark on 9 April 1940, the Althing replaced the King with a regent and declared that the Icelandic government would take control of its own defence and foreign affairs. &
With Elizabeth's accession, it seemed probable the royal house would bear the Duke of Edinburgh's name, in line with the custom of a wife taking her husband's surname on marriage. &
-0:30 \\
& A month later, British armed forces conducted Operation Fork, the invasion and occupation of the country, violating Icelandic neutrality. &
The British Prime Minister, Winston Churchill, and Elizabeth's grandmother, Queen Mary, favoured the retention of the House of Windsor, and so on 9 April 1952 Elizabeth issued a declaration that Windsor would continue to be the name of the royal house. & 
-0:30 \\
\bottomrule
\end{tabular}
\end{scriptsize}
\caption{Claims and evidence shown to players in the voting phase: the voter must detect which claim is incorrect.  Initially, the player only sees the claim---if the player can answer with only that, they get the most points.  Automatically found evidence is shown one by one upon the voter's request.  Waiting and asking for evidence both decrease the time---and the points---available.  Eventually, if time does not run out, the gold evidence selected by the author of the claim is shown.}
\flushright \rotatebox{180}{Answer: The claim about Elizabeth II is refuted by the evidence.}
\label{tab:claims}
\end{table*}

\subsection{Incentive Structure}

Players earn points in two ways: either spotting incorrect claims by
voting as early as possible or authoring challenging claims.
They alternate between the two roles in every game session.
These two rewards are in opposition to each other.
%

Because the goal of the voters is to find the claim that is
incorrect, claim authors (of either entailed or refuted claims) only get
points when voters are \emph{not} fooled {\bf and} when the voters need evidence.
The total points are split between the voter and
authors when the voter correctly guesses, making
this a \emph{zero-sum} game.
As a voter requests evidence or takes more time, a larger fraction of
the total points will go to authors.
Thus, authors are encouraged to write difficult claims; voters are
encouraged to select claims correctly.

When a voter guesses incorrectly, they get no points, to ensure the
examples are valid.
While incorrect guesses can happen for impossible
claims, writing claims that are merely difficult is a better
strategy since easy claims that may be spotted quickly are awarded no
points.\footnote{We also allow players to flag obscene, incorrect, or
  otherwise problematic claims.}

In addition to humans voting on claims, we also ask users which of
the two claims they ``like'' more, independent of voters' accuracy.
People like true claims (0.39) more than false claims (0.35, $t=2.53$,
$p=0.01$), except for claims about science and technology, where
people prefer false claims (0.46) more than true claims (0.32,
$t=-2.50$, $p=0.02$).
Authors get points when voters like their claims; this
additional incentive encourages authors to create interesting \emph{and}
surprising examples.

\section{Methods: Subtasks and Models}
\label{sec:models}

Each of the instances in \bigfmt{} is a tuple $(c, e, l)$:
a natural language claim~$c$,
evidence~$e$ from a knowledge corpus $\mathcal{K}$
(in our case Wikipedia),
and a binary label~$l$ (entailment / contradiction).\footnote{Unlike
  \fever, we do not allow authors to write claims that lack
  ``enough information''.}
From this we define two sub-tasks, following~\citet{fever-18}. 
The first sub-task, retrieval, requires
systems to select candidate evidence from $\mathcal{K}$
(including, perhaps, the gold evidence~$e$).
The second sub-task is entailment, where systems
given claim~$c$ and the gold evidence~$e$
need to make a final prediction for the label~$l$.
We also consider an end-to-end setting.  Instead of the gold evidence,
systems only have access to the retrieved evidence $\hat{e}$ at test
time.
In the rest of this section we define baseline models for each of the
sub-tasks.

\subsection{Retrieval}
\label{sec:ir}
Our setting resembles the retrieval setting in
the \abr{kilt} benchmark~\cite{petroni2020kilt}, but the results are
evaluated at the evidence level as opposed to the page level, to
represent a more realistic use case.
The evidence corpus can be found online\footnote{\href{https://github.com/facebookresearch/KILT/tree/master/kilt/retrievers}{http://github.com/facebookresearch/KILT/}}
and consists of twenty-two million text passages, each having a length of
a hundred words, from five million pages of the English Wikipedia image from
August 2019.
We align gold \bigfmt{} evidence to this knowledge source by
selecting the passage with highest overlap with each evidence
sentence, according to the modified \mbox{$n$-gram} precision component of the
\abr{bleu}~\cite{papineni-02}.
We remove \num{1598} examples\footnote{
This happens because \bigfmt{} was constructed from a more recent version of
Wikipedia than \abr{kilt}.
} where the precision was less than $0.5$.

We evaluate two baselines. The first one follows
\citet{chen-etal-2017-reading} and uses a \abr{tf-idf} retrieval model
with unigrams and bigrams and $2^{20}$ hash buckets.
The title of page is added to the passage content for additional
context.
The second baseline uses \emph{Dense Passage
  Retrieval}~\citep[\abr{dpr}]{karpukhin2020dense}, using the same
fixed pre-trained passage embeddings and query encoder as the ones used in
\citet{petroni2020kilt}.

\subsection{Entailment}

For the second component of the task, we follow state-of-the-art
entailment models~\cite{zhou-etal-2019-gear, liu-etal-2020-fine,
  eisenschlos-etal-2020-understanding}: given the concatenated gold evidence and claim, a \abr{bert}-base model~\cite{devlin-19} outputs a binary entailment / contradiction label.

For end-to-end label accuracy, we use the same models but test only retrieved (rather than gold) passages. During training we include both the gold and the top two retrieved passages.


\section{Experiment Results: Machines Spotting False Claims}
\label{sec:results}

This section studies the performance of existing automatic methods on \fmt{}
for both the retrieval of evidence (Section~\ref{sec:retrieval-results}) and for entailment once the results are retrieved (Section~\ref{entailment-results}).


\subsection{Retrieval Results}
\label{sec:retrieval-results}

Retrieving evidence for \bigfmt{} is considerably harder
(Table~\ref{tab:retrieval}); we also include comparable results on \fever{}. 
The documents retrieved by \abr{dpr} are consistently better than the
ones by a \abr{tf-idf} system for both of the datasets we tested,
which is consistent with other work on dense text retrieval
\cite{realm}.

\begin{table}
\begin{center}
\scalebox{0.65}{
\begin{tabular}{llccc}
\toprule
   & Dataset & R-Precision & Recall@5 & Recall@10 \\
\midrule
  \multirow{2}{*}{TF-IDF}  & \fever  & 25.3  & 44.1  & 53.2 \\
                           & \bigfmt & 10.4  & 21.2  & 28.3 \\
\midrule
  \multirow{2}{*}{DPR}     & \fever  & 32.0  & 50.4  & 58.7 \\
                           & \bigfmt & 25.3  & 42.6  & 51.0 \\
\bottomrule
\end{tabular}
}
\end{center}
\caption{Results of evidence retrieval baselines on \bigfmt{} and
  \fever. \emph{R-Precision} is defined as the precision@$k$, where $k$ is the
  number of gold evidence snippets for the claim. \bigfmt{} is harder
  for both the retriever systems.}
\label{tab:retrieval}
\end{table}

\subsection{Entailment Results}
\label{entailment-results}

This section presents the results of training a
\abr{bert}~\cite{devlin-19} model for the entailment task of
\bigfmt{}. Given a claim and the gold evidence, does the evidence
support or refute the claim?
To compare with \fever, we discard all
\emph{not enough evidence} examples, because the lack of
evidence for this class makes it trivial to classify correctly.
%

Following ~\citet{gururangan-etal-2018-annotation}, we first train a
claim-only classifier, which ignores the evidence text.
\bigfmt{} examples are harder to classify without looking at the
evidence (Table~\ref{tab:entailment}), indicating that the claims
contain fewer ``give away'' artifacts compared to \fever{} as already
suggested by Table~\ref{tab:all_artefacts}.  We provide additional discussion in Appendix~\ref{sec:lmi}.

%


Like the techniques proposed by \citet{clark-etal-2019-dont}, the
claim-only classifier can also be used on both \bigfmt~ and \fever~ to
split the dev sets into ``easy'' and ``hard'' partitions:
The \easy~ partition contains all examples correctly classified
by a claim-only classifier, and the \hard~ partition has everything else.
The similar accuracy of the \bigfmt{} dev and \hardfever{} dev partitions further suggests that \bigfmt{} is comparable
to the harder and higher-quality subset of \fever~ (Table~\ref{tab:entailment}).

\begin{table}
\begin{center}
\scalebox{0.8}{
\begin{tabular}{lcccc}
\toprule
Dataset & Claim-Only & \easy & \hard & \textsc{all} \\
\midrule
\bigfmt & 61.9       & 86.1  & 66.4  & 78.1 \\
\fever  & 79.1       & 97.1  & 79.3  & 93.3 \\
\bottomrule
\end{tabular}
}
\end{center}
\caption{Comparison of dev accuracy between \fever{} and \bigfmt{} for
  different partitions of the data and when using only claims.
The partition into \easy~ and \hard~ splits is based on the claim-only
classifier: the claim-only classifier can solve \easy{} examples.
\bigfmt~ examples thus are comparable to \hard{} \fever{}'s difficulty.
}
\label{tab:entailment}
\end{table}


We also train an end-to-end verification model that, rather than taking evidence as given, must use noisy passages from a retrieval system (Section~\ref{sec:ir}).
At train time, we generate multiple training instances for each claim
using either the gold evidence or the top two retrieved examples.
At prediction time, we average the logit scores of each of the top-$k$
retrieved passages (Table~\ref{tab:e2e}). We
include
a so-called \emph{oracle} setting for a fair comparison of the
improvement margin. This number differs from
Table~\ref{tab:entailment} in that it uses a single gold $100$ word
passage as evidence instead of short sentences, which might introduce noise.

\begin{table}
\begin{center}
\scalebox{0.9}{
\begin{tabular}{lccc}
\toprule
  Retriever & Top-1 & Top-3 & Top-5 \\
\midrule
  Oracle & 69.3 & -- & -- \\

\midrule
  \abr{tf-idf} & 62.3  & 62.0  & 61.2 \\
  \abr{dpr}    & \textbf{64.2}  & 63.6  & 63.9 \\
\bottomrule
\end{tabular}
}
\end{center}
\caption{End-to-end label accuracy results of retrieval followed by
  entailment on \bigfmt{}.  We vary the number of retrieved examples at
  prediction time. We compare against using the gold evidence as an
  oracle, which differs from Table~\ref{tab:retrieval} in using a
  single $100$ word passage as evidence.}
\label{tab:e2e}
\end{table}

\begin{table*}[t]
\small
\centering
\begin{tabular}{@{}cccl@{}}
\toprule
Name      & Ratio & Label & Claim \& Gold Evidence                                                                                                                                                                                                                                                                                                                                                                                  \\ \midrule
Temporal  & 26\%  & R     & \begin{tabular}[c]{@{}l@{}}\underline{Claim}: The Flavian Amphitheatre, which was mainly used for gladiatorial contests,\\ could hold over 50,000 people, and \textbf{animal hunts continued until the 10th century}.\\ \underline{Evidence}: \textbf{Animal hunts continued until at least 523}, when Anicius Maximus celebrated\\ his consulship with some venationes, criticised by King Theodoric for their high cost.\end{tabular} 
\\ \midrule
Reasoning & 26\%  & S     & \begin{tabular}[c]{@{}l@{}}\underline{Claim}: Darius Milhaud was a French composer that had a child that was \textbf{his second Cousin}.\\ \underline{Evidence}: In 1925, Milhaud \textbf{married his cousin}, Madeleine (1902–2008), an actress and \\reciter. In 1930 she gave birth to a son, the painter and sculptor Daniel Milhaud, who was\\ the couple's only child.\end{tabular}
\\ \midrule
Paraphrase  & 22\%  & R     & \begin{tabular}[c]{@{}l@{}}\underline{Claim}: Sister Carrie sold poorly, and was criticized for taking the \textbf{Lord's title in vain}.\\ \underline{Evidence}: The book was also criticized for \textbf{never mentioning the name of God}.\end{tabular}
\\ \midrule
Diversion   & 16\%  & S     & \begin{tabular}[c]{@{}l@{}}\underline{Claim}: Following his retirement from the MLB, Prince Hal became a \textbf{top executive} of\\ a \textbf{real estate} company.\\ \underline{Evidence}: After his retirement from baseball, Newhouser was away from the sport for\\ 20 years, serving as a \textbf{bank vice president}.\end{tabular}
\\ \midrule
Controversy & 8\%   & S     & \begin{tabular}[c]{@{}l@{}}\underline{Claim}: Francis Marion fought in the Revolutionary War and was an influence for the\\ protagonist in the movie, The Patriot, where \textbf{his character highly altered to show him as}\\ \textbf{good natured}.\\ \underline{Evidence}: Sean Busick \dots says that based on the facts, ``Marion deserves to be remembered\\ as one of the heroes of the War for Independence.'' \dots the film's depiction of Martin ``as a\\ \textbf{family man and hero} who single-handedly defeats countless hostile Brits'' \dots was one of\\ the ``egregious oversights'' that TIME magazine cited when listing \textit{The Patriot} as\\ number one \dots \textbf{historically misleading} [film]''\end{tabular} 
\\ \bottomrule
\end{tabular}
    \caption{An ontology of human strategies for creating challenging claims in our dataset, sampled from claims that challenged both humans and computers.}
    \label{tab:ontology}
\end{table*}

\section{Dataset Analysis: Humans Spotting and Writing False Claims}
\label{sec:analysis}

While the previous section focuses on how well automatic methods can detect false claims, this section focuses on human ability.
Voters are usually right and were fooled $20.40\%$ of the time. 
This section addresses how players are fooled and how this compares to computers.

To provide a better picture of the strategies players use to craft challenging claims, we manually sample fifty instances from the development set that \emph{both} models and humans answer incorrectly. 
We focus on these examples because they are the most difficult and are the emphasis of our adversarial technique.
Two claims were mislabeled and two more lacked a necessary evidence span.
Table~\ref{tab:ontology} shows examples of each of the strategies, which we discuss in more detail in this section.

\paragraph{Temporal}

Many of the most challenging claims require an inference about time: whether one event happened before another, how long an event happened, or whether an event happened during a period.
While many of these are based on years, centuries, or other explicit markers of time, some authors use narrative time.
For example, the page for the novel \textit{As I Lay Dying} describes the plot in order, so it's difficult for either a system or a human given sentences (without knowing where they appear in the original page) to know when Addie Bundren dies.
This shows some of the limitations of the setup: not only must voters reason across multiple pieces of evidence, this reasoning is only possible if they know the \emph{order} in the underlying evidence.
Other markers of time include ``the pilot'' for the first episode of \underline{The Office}; readers must realize that if Kelly Kapoor was introduced in the episode \textit{Diversity Day}, that implies Mindy Kaling's character did not appear in the pilot.

\paragraph{Reasoning}

A related, but more general, strategy requires the reader to reason: mathematically, applying definitions, or understanding hyponomy.
For example, knowing that the child of your cousin is your second cousin or recognizing that ``This mirrors the Disney Parks East regional division consisting of Shanghai Disney Resort, Hong Kong Disneyland and Walt Disney Attractions Japan\dots'' implies that there are more than two Walt Disney resorts outside of the United States.

\paragraph{Paraphrase} 

A well-known strategy to confuse entailment systems is to change words so that there are fewer exact matches.  
Some of these are straightforward: ``Titration is used when doctors test how much sugar is in a patient's liquid waste'' is almost a direct paraphrase of ``glucose in urine may indicate diabetes in a patient''.  
Other paraphrases are more poetic: ``Charles Evans Hughes shuffled off this mortal coil in Massachusetts, and then was taken to New York to be submerged in soil'' paraphrasing ``Hughes died in what is now the Tiffany Cottage of the Wianno Club in Osterville, Massachusetts. He is interred at Woodlawn Cemetery in the Bronx, New York City''.
These paraphrases are realistic, similar to how humans might restate facts to make them more accessible or more interesting to a reader.

\paragraph{Diversion}

An interesting strategy to fool the retrieval phase of \textsc{fever} systems is to create claims that point to specific text \emph{but not the text that refutes or supports the claim}.
For example, ``Following his retirement from the \abr{mlb}, Prince Hal became a top executive of a company'' retrieves information about how Hal Newhouser earned the nickname ``Prince Hal'' and his later business investments but not his post-baseball career in banking.

\paragraph{Controversy}

A more fundamental issue with entailment systems is that even trusted sources such as Wikipedia contain contradictory evidence.
This is most prominent 
with interpretations of works of fiction, where there are multiple theories about the same work.
A skillfully written claim can retrieve one viewpoint while using an opposing viewpoint as the gold evidence.

For example, one claim strongly took the position that the end of the film \textit{Inception} was a dream.
Voters saw evidence to the contrary and thought the claim was refuted.
Because systems focus on the highest scoring retrieved passages (as do the human voters), this lead both humans and computers to overlook the disputed interpretations.

\subsection{What was difficult for humans?}

The amount of evidence a human needs is a unique metric of how difficult a claim is for humans (although incremental evidence is recommended for question answering systems in \newcite{boyd-graber-20}, to the best of our knowledge it has not been applied to entailment or validation).
The claims that most challenge humans typically use \emph{diversion} (e.g., ``\textit{The Quiet Man} was a song by Bing Crosby about a soldier who lost his voice from a bomb in World War 2''), which is particularly challenging for retrieval systems.
Other common strategies for the claims most challenging for humans were  \emph{paraphrase}, which can ``hide'' the relevant evidence and prevent retrieval, and \emph{reasoning}, which often requires multiple pieces of evidence to reach a conclusion.
\section{Limitations and Conclusion}
\label{sec:conc}

While this paper seeks to advance the ability of humans and computers to support or refute statements entailed from a static, reliable source, the goal of examining arbitrary statements remains elusive.
By construction, we have focused on statements that are incorrect because of \emph{factual} errors.
Other datasets that use human-sourced obfuscations or deception are more nuanced and use framing or shading~\cite{pan-93}, which models trained on this dataset cannot detect.
Our goal is to focus on clear facts that can be recognized by computers, which is already challenging enough.

Further improving verification likely requires creating targeted datasets that focus on specific strategies for creating statements that are refuted by evidence, perhaps selecting different explanations for particular users~\cite{feng-19}.
Likewise, a more complicated task likely requires more nuanced incentives and instructions for authors.
However, this dataset provides a foundation to build these richer, more challenging datasets for entailment.


\section*{Ethical Considerations}
As our work involves human participants, all players provided informed consent and no personally identifiable information (\abr{pii}) was collected or will be released. The collected data have been vetted for presence of \abr{pii} as well as offensive language through heuristics and random sampling. 

Some participants received fair compensation in the United States in exchange for playing the game, but that compensation was not tied to speed or accuracy to prevent distorting the motivation of players. Intrinsic motivation, such as curiosity, competitiveness, creative drive and fun, rather than extrinsic motivation has been shown to produce higher quality results~\cite{payenough}. 

The released data and the experiments we conducted are in
English, therefore we do not claim generalization of our findings across languages. However, we believe that the proposed methods could be
applied in other languages using other available corpora as a source of evidence.

\section*{Acknowledgements}

First and foremost, we would like to specially thank Connie Tao for her guidance and assitance in managing the project.  The project would also have been impossible without the \fmt{} players. We also would want to thank Thomas M{\"u}ller, William Cohen, Dipanjan Das, Slav Petrov, Pedro Rodriguez, Massimiliano Ciaramita, and Christian Buck for comments on the drafts and testing the game.  We also thank 
the anonymous reviewers for their time, constructive
feedback, useful comments and suggestions about this work.  Boyd-Graber is supported by \abr{nsf} Grant \abr{IIS}-1822494.

\bibliographystyle{style/acl_natbib}
\bibliography{bib/journal-full,bib/jbg}

\clearpage

\appendix\section*{Appendix}

\section{Experimental Setup}

In this section we provide details on the hyper-parameters used and the experimental setup. All BERT models described are of base size ($12$ layers, $16$ attention heads, $768$ hidden dimension), and contain $110$ million parameters.

The training is done for $10$ epochs, a learning rate of $10^{-5}$. We use a batch size of $32$ and a learning rate of $512$. On a single Cloud TPU v2 the model can process one batch $180$ms, and a full epoch in around one minute. For all the reported results we take the median over $3$ random seeds.

\section{Game Interface}
\label{app:game}
\begin{figure*}
\centering
\includegraphics[width=0.9\textwidth]{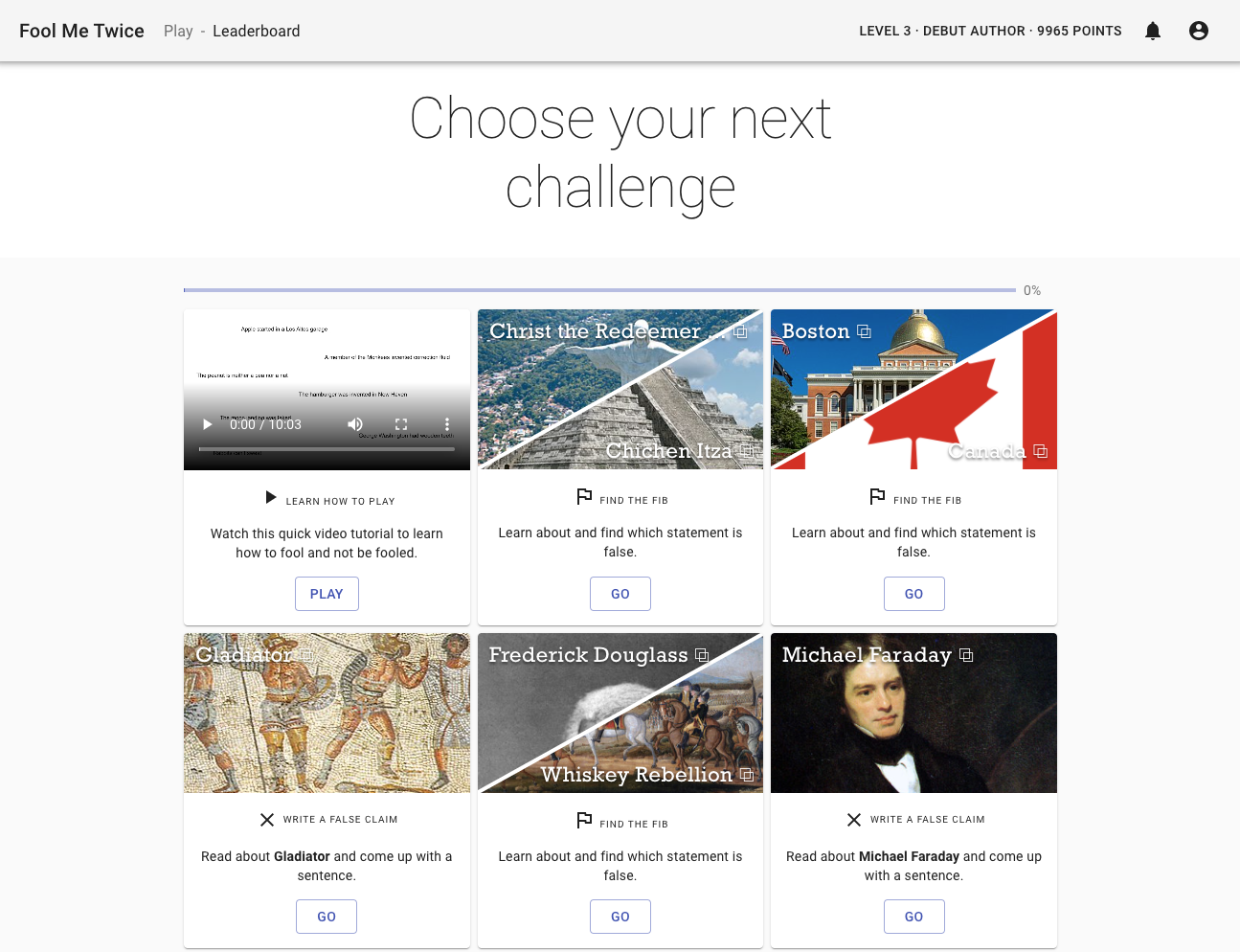}
  \caption{Menu where players select between authoring or voting on claims. A diverse set of categories is presented to engage people according to their interests.}
  \label{fig:ui-menu}
\end{figure*}

\begin{figure*}
\centering
\includegraphics[width=0.9\textwidth]{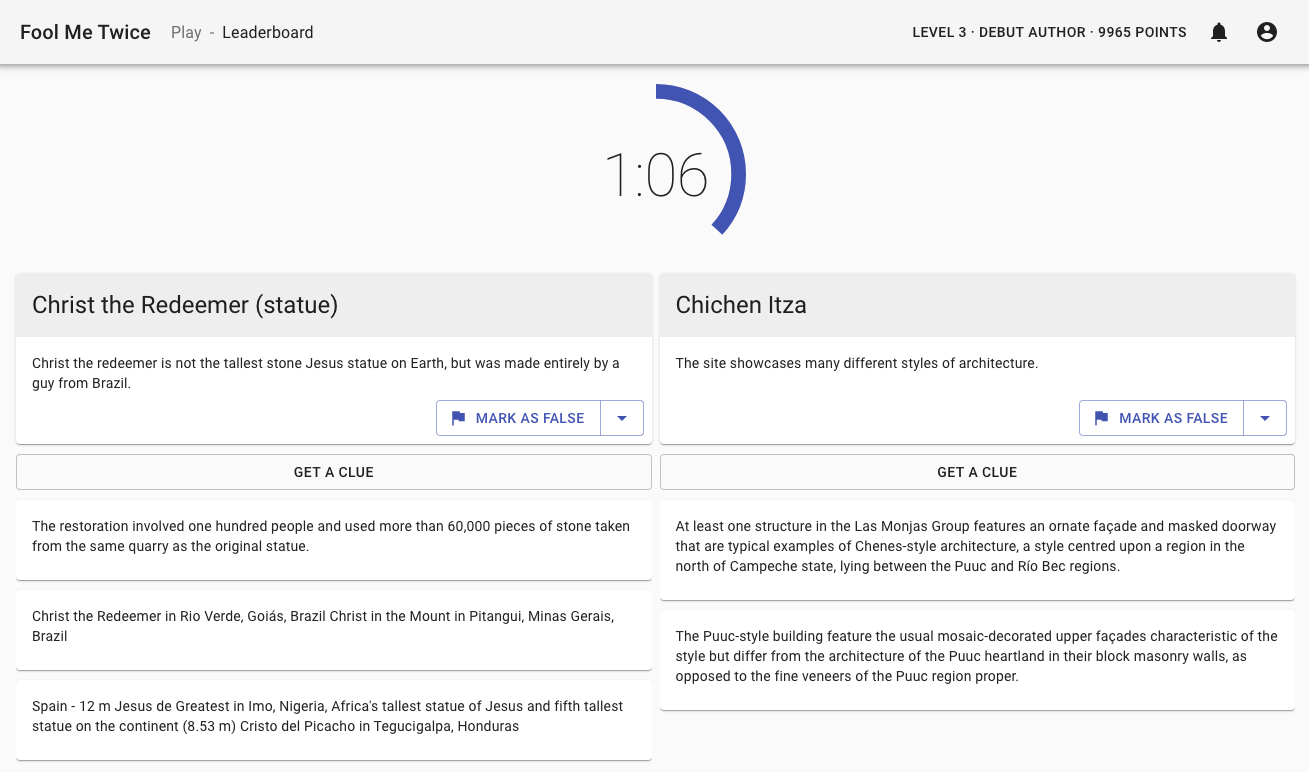}
  \caption{The voting interface shows one entailed and refuted claim. The player has two decide which one is the refuted one before time runs out. Getting clues consumes 30 seconds in the timer.}
  \label{fig:ui-vote}
\end{figure*}

\begin{figure*}
\centering
\includegraphics[width=0.9\textwidth]{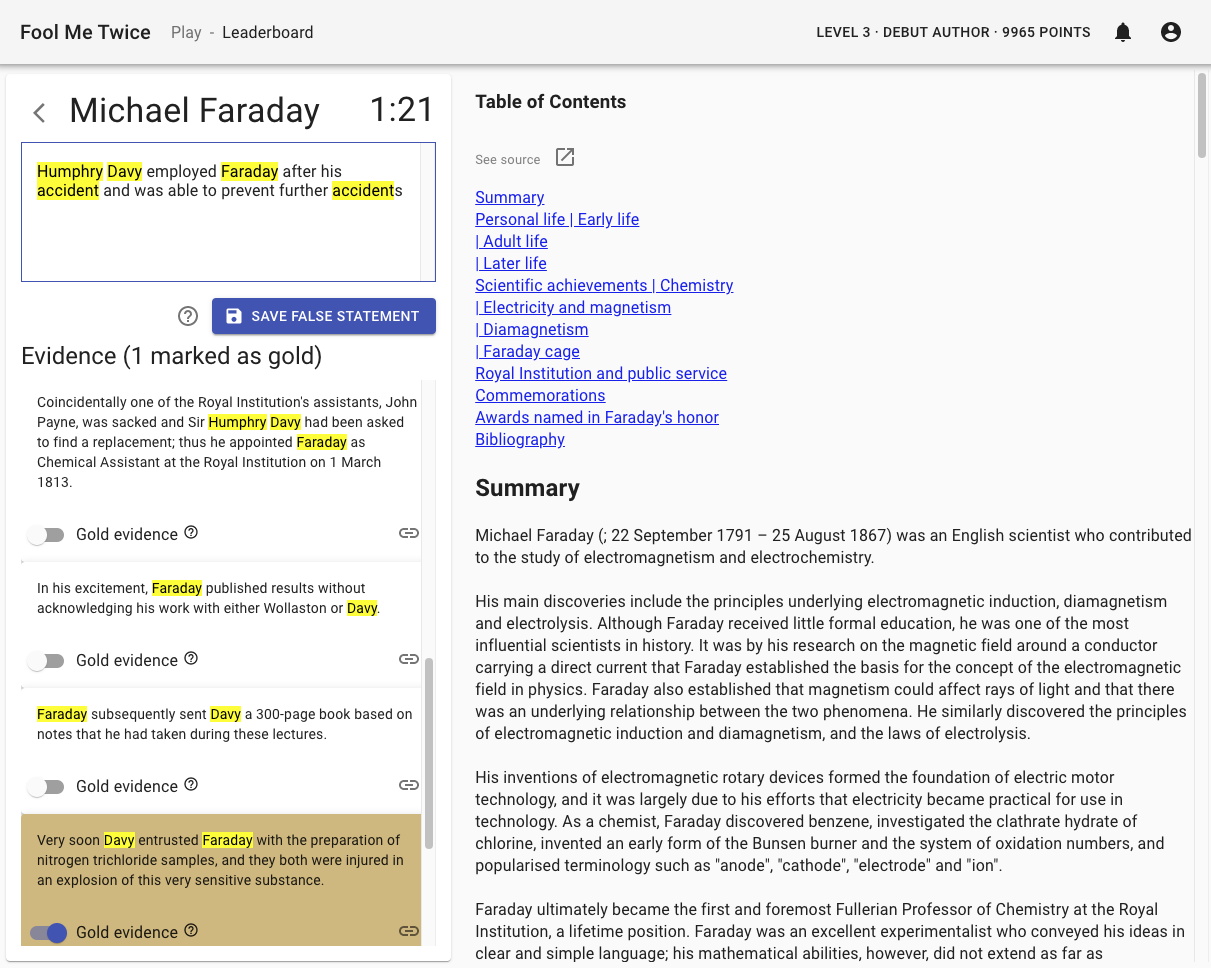}
  \caption{In the writing screen, players are asked to write either an entailed or refuted evidence given the evidence on the right hand side. As they write, a retrieval system picks the most relevant evidence. They can mark the gold evidence that supports or contradicts the claim, and are instructed to write in such a way that the gold evidence is not at the top of the retrieved list.}
  \label{fig:ui-write}
\end{figure*}

In this section we include screenshots of the three main screens of the game. Figure~\ref{fig:ui-menu} shows menu interface that allows players to choose topics according to their interests, we include many Wikipedia categories to ensure a diverse set of options. Figure \ref{fig:ui-vote} has example of the voting game, the simplest and fastest way to engage with the game and understand how to be a good author as well. Finally, figure \ref{fig:ui-write} shows the authoring user interface, that displays the retrieved and selected gold evidence as the user types. Matching tokens in the text and the retrieved evidence are highlighted.

\section{Local Mutual Information}
\label{sec:lmi}
Tables~\ref{tab:fever_artefacts}, \ref{tab:fm2_artefacts} list the top-10 predictive bigrams for the \textsc{REFUTES} label using ~\citet{schuster-etal-2019-towards}'s method of computing \emph{Local Mutual Information} (\abr{lmi}), defined for a bigram $b$ and label $l$ as:  
\[
\abr{LMI}\left(b, l\right) = p\left(b, l\right)\cdot \log\left(\frac{p\left(l\mid b\right)}{p\left(l\right)}\right)
\]
where the probabilities use the empirical counts.

Consistent with the much lower claim-only classifier (see Table~\ref{tab:entailment}), \bigfmt~ contains no ``give away'' bigrams that are highly predictive of the label on both the training and development data whereas, as previously reported by ~\citet{schuster-etal-2019-towards}, \fever~ has many.  Moreover, the ``quality'' of predictive bigrams for \fever~ suggests that annotators (subconsciously) used specific strategies when writing \textsc{refutes} examples (``is only'', ``did not'', ``is incapable''), but no such patterns can be seen for \bigfmt.

\begin{table}
    \centering
    \begin{tabular}{lp{1.8cm}p{1.8cm}}
    \toprule
        Bigram  &     Train LMI~$\times 10^-5$   &      Dev LMI~$\times 10^-5$ \\
    \midrule
        is only & $622$ & $938$ \\
        did not & $859$ & $528$ \\
        not a & $775$ & $481$ \\
        was not & $729$ & $-$ \\
        incapable of & $721$ & $710$ \\
        only a & $455$ & $717$ \\
        is incapable & $474$ & $551$ \\
        was only & $-$ & $536$ \\
        has only & $447$ & $-$ \\
        yet to & $420$ & $384$ \\
        of being & $-$ & $385$ \\
    \bottomrule
    \end{tabular}
    \caption{Top-10 highest \abr{lmi} bigrams for \textsc{REFUTES} label in \fever~ for both Train and Dev.  Note the large overlap of label-predictive bigram artefacts.}
    \label{tab:fever_artefacts}
\end{table}

\begin{table}
    \centering
    \begin{tabular}{lp{1.8cm}p{1.8cm}}
    \toprule
        Bigram  &     Train LMI~$\times 10^-5$   &      Dev LMI~$\times 10^-5$ \\
    \midrule
        by a & $-$ & $562$ \\
        mad , & $-$ & $502$ \\
        , mad & $-$ & $502$ \\
        on the & $-$ & $473$ \\
        innocent iii & $-$ & $467$ \\
        statue of & $-$ & $426$ \\
        for his & $-$ & $407$ \\
        pope innocent & $-$ & $407$ \\
        mary , & $-$ & $365$ \\
        queen of & $-$ & $365$ \\
        the second & $338$ & $-$ \\
        is a & $312$ & $-$ \\
        was a & $307$ & $-$ \\
        was the & $306$ & $-$ \\
        is the & $233$ & $-$ \\
        of his & $200$ & $-$ \\
        has never & $189$ & $-$ \\
        was born & $177$ & $-$ \\
        written by & $165$ & $-$ \\
        about a & $162$ & $-$ \\
        \bottomrule
    \end{tabular}
    \caption{Top-10 \abr{lmi} bigrams for \textsc{REFUTES} label in \bigfmt~ for both Train and Dev.  Note both the absence of ``give-away'' bigram overlap beween Train and Dev;  and the more ``random'' quality of predictive bigrams compared to those for \fever~ in Table~\ref{tab:fever_artefacts}}
    \label{tab:fm2_artefacts}
\end{table}

\end{document}